\documentclass{article}

\usepackage[nonatbib, preprint]{neurips_2023}

\usepackage[utf8]{inputenc} 
\usepackage[T1]{fontenc}    
\usepackage{hyperref}       
\usepackage{url}            
\usepackage{booktabs}       
\usepackage{amsfonts}       
\usepackage{nicefrac}       
\usepackage{microtype}      
\usepackage{xcolor}         
\usepackage{graphicx}
\usepackage{enumerate}
\usepackage{multirow}
\usepackage{soul} 
\usepackage{color}
\definecolor{Gray}{gray}{0.9}
\definecolor{Gray2}{gray}{0.95}
\usepackage{color,colortbl}
\usepackage{amstext}
\usepackage{subcaption}
\usepackage{biblatex} 
\title{SRMAE: Masked Image Modeling for Scale-Invariant Deep Representations}

\author{
  Zhiming Wang \\
  Beihang University \\
  \texttt{2306418@buaa.edu.cn} \\
  \and
  \textbf{Lin Gu} \\
  The University of Tokyo, Japan \\
  \texttt{lin.gu@riken.jp}
  \and
  \textbf{Feng Lu} \\
  Beihang University \\
  \texttt{lufeng@buaa.edu.cn} \\
}

\begin{document}

\maketitle

\begin{abstract}
Due to the prevalence of scale variance in nature images, we propose to use image scale as a self-supervised signal for Masked Image Modeling (MIM). Our method involves selecting random patches from the input image and downsampling them to a low-resolution format. Our framework utilizes the latest advances in super-resolution (SR) to design the prediction head, which reconstructs the input from low-resolution clues and other patches. After 400 epochs of pre-training, our Super Resolution Masked Autoencoders  (SRMAE)  get an accuracy of 82.1\% on the ImageNet-1K task. Image scale signal also allows our  SRMAE to capture scale invariance representation. For the very low resolution (VLR) recognition task, our model achieves the best performance, surpassing DeriveNet by 1.3\%. Our method also achieves an accuracy of 74.84\% on the task of recognizing low-resolution facial expressions, surpassing the current state-of-the-art FMD by 9.48\%.

\end{abstract}

\section{Introduction}
Masked Image Modeling (MIM)\cite{beit, MAE, SimMIM} self-supervisely learns deep representations by masking a portion of input signals and predicting these masked signals. MIM has shown has shown great success in  downstream tasks such as image classification \cite{imagenet1k}, object detection \cite{coco}, semantic segmentation \cite{ADE20K}, and video classification \cite{videodatabase}.



The MIM method is a form of self-supervised learning that utilizes signals during the learning process. Researchers are currently exploring the use of multiple self-supervised signals for learning through a detailed analysis of the MIM method. MAE consider the original pixel intensity as a self-supervised signal by reconstructing it based on partial observation. MaskFeat demonstrated that masked features like graident histogram is a powerful self-supervised signal. CAEv2 utilizes CLIP as a self-supervised signal.


In our study, we propose using scale variance as a self-supervised signal. Scale variance is prevalent in nature images due to the continuous variation of the projection size of an object relative to the human eye or camera sensor as its distance changes~\cite{scale-invariant}. This scale variance poses numerous challenges for modern neural networks ~\cite{azulay2018deep}. A minute variation in size that is imperceptible to the human eye can cause a significant change in the output. Thus, scale invariance is an integral aspect of the advancement in computer vision\cite{lindeberg2012scale}. Recent studies have demonstrated that resolution/scale is an efficient self-supervised signal that enables the network to generalize its object detection capability to low-resolution (LR) images.

\begin{figure}[t]
\begin{center}
   \includegraphics[width=0.8\linewidth]{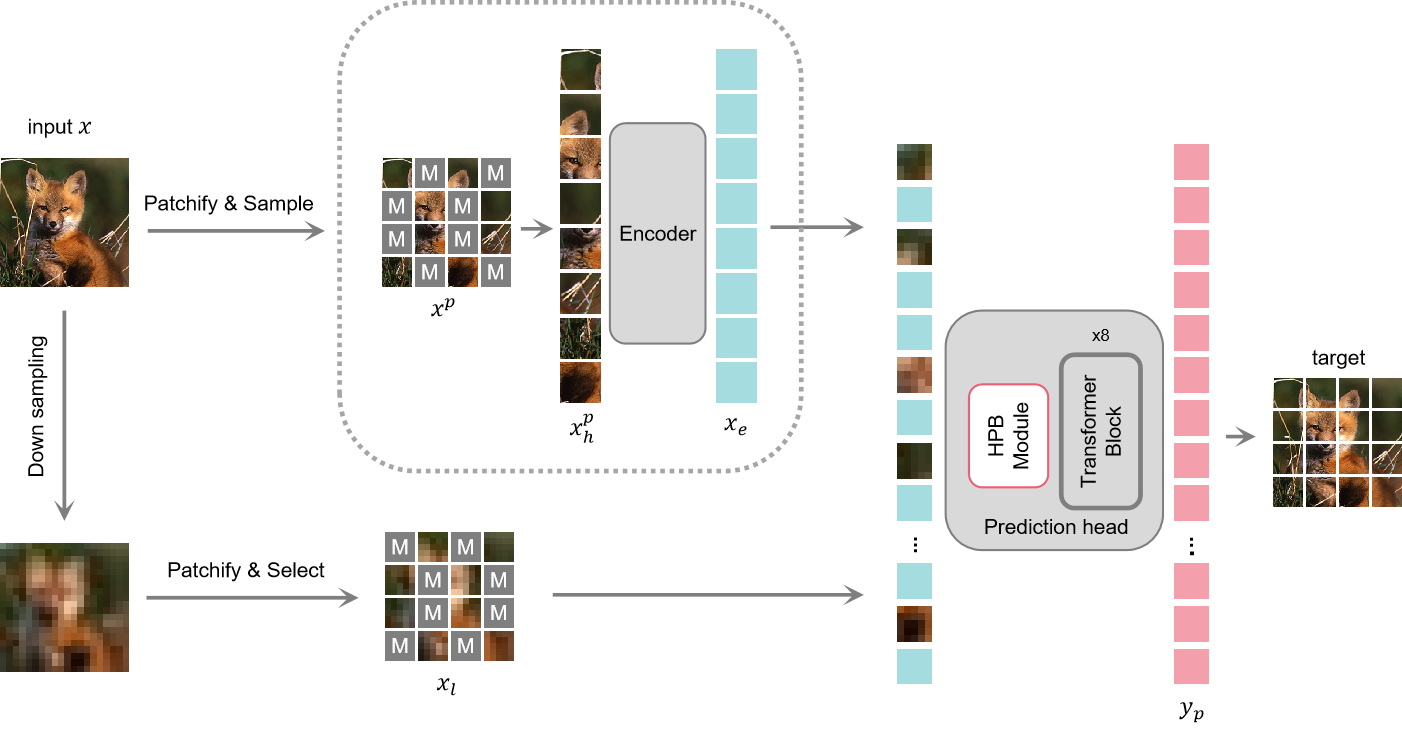}
\end{center}
    \caption{SRMAE Framework: we randomly sample the input image and downsample the low-resolution patches. Subsequently, we use an encoder to obtain the latent representation of the unsampled patches. Our prediction head restores the original resolution using the remaining patches and the LR signals. The gray box section has a consistent encoder design with MAE.}
\label{fig:SRMAE_model}
\end{figure}

We introduce scale as a self-supervised signal in our proposed novel MIM algorithm called Super Resolution Masked Autoencoders (SRMAE). Fig. \ref{fig:SRMAE_model} illustrates that the input image is divided into a set of patches, some of which are downsampled into low-resolution (LR) patches. On the other hand, the remaining patches are called high-resolution (HR) ones, similar to the MAE \cite{MAE}, where our SRMAE also encodes the HR patches. In contrast to MAE, which solely reconstructs the input image from these HR patches, we use both LR and encoded HR patches to reconstruct the original signal. The empty mask token of MAE is substituted with the LR signal in the prediction head. Additionally, we employ a High Preserving Block (HPB) module \cite{ESRT} to extract features from LR patches before reconstructing the original signal with a lightweight ViT (Vision Transformers) structure to benefit from the super-resolution (SR) technique.


The original image is termed as high-resolution (HR) image. Similar to the ViT \cite{ViT}, the high-resolution image and the low-resolution image are divided into two sets of regular non-overlapping patches. We sample a subset of patches from the high-resolution patches and take the low-resolution patches that are relative to the unsampled high-resolution patches as the subsequent prediction head input. The encoder of SRMAE, similar to the encoder of MAE \cite{MAE}, processes only these sampled high-resolution patches. Nevertheless, unlike the prediction head of MAE, we preserve the low-resolution signals of patches that have not been encoded. The complete set of tokens comprises encoded high-resolution patches and selected low-resolution patches that act as the input to the SRMAE prediction head. In contrast, the full set of tokens for MAE consists of encoded visible patches and mask tokens. We add positional embeddings to all full set tokens, and the prediction head restores the resolution for the selected low-resolution patches. Rather than reconstructing pixels like MAE, we need to recover the resolution. To achieve this goal, we use an HPB module \cite{ESRT} that is appropriate for extracting features before the lightweight ViT structure. In the fine-tuning stage of ImageNet-1K, SRMAE with ViT-B achieved a top-1 accuracy of 82.1\%, which is 1\% lower than MAE with an equal number of training epochs.

The use of self-supervision with the scale signal makes our SRMAE capable of capturing scale-invariant representations. (Very) low resolution (VLR/LR) recognition is particularly challenging as neuron networks have to recognise under various scales. We thus evaluated the performance of our SRMAE on several VLR/LR tasks. In particular, we conducted a VLR digit classification experiment on the SVHN dataset using images of resolution 8$\times$8. We conducted LR facial expression classification experiments using images of resolution 100$\times$100 on the CK+, RAF-DB, and ExpW datasets. Our SRMAE shows strong scale-invariant ability when compared with state-of-the-art (SOTA) methods such as DeriveNet\cite{DeriveNet}, which obtained a top-1 classification accuracy of 87.85\% on the SVHN dataset, and FMD\cite{FMD}, which achieved 65.37\% top-1 accuracy on the ExpW dataset. Our SRMAE has achieved SOTA performance on some VLR/LR recognition datasets. On the SVHN dataset, our SRMAE achieved a top-1 accuracy of 89.14\%, which is 1.3\% higher than the previous state-of-the-art (87.85\%\cite{DeriveNet}). On the ExpW dataset, our SRMAE achieved a top-1 accuracy of 74.84\%, which is a significant improvement of 9.5\% over the previous state-of-the-art top-1 accuracy of 65.37\%.


In this paper, we mainly make the following contributions:
\begin{enumerate}[(i)]
    \item For the first time, we propose to utilise scale as a self-supervised signal for Masked Image Modeling (MIM). Our SRMAE is a simple and easy-to-implement framework preserving scale invariance. 
    \item Our framework could leverage the latest super-resolution (SR) architecture advance to design the prediction head that predicts masked original signals from low-resolution clues.
    \item SRMAE achieves close to SOTA results in different resolution visual tasks, such as fine-tuning on ImageNet-1K, VLR digit classification on SVHN dataset and LR facial expression classification on ExpW dataset.
\end{enumerate}

\section{Related Work}
\noindent\textbf{Masked Image Modeling}
Masked Image Modeling (MIM) has become a popular pretext task for visual representation learning, inspired by BERT\cite{bert} for Masked Language Modeling\cite{beit,MAE,SimMIM}. The context encoder approach\cite{Contextencoders} is a groundbreaking work in this field, which masks a rectangular section of the original images and predicts the missing pixels. iGPT\cite{chen2020generative}, ViT\cite{ViT}, and BEiT\cite{beit} were the first notable works that recognized the MIM learning strategy on modern vision Transformers. Several target signals have been developed for the mask-prediction pretext task in MIM, such as normalized pixels\cite{MAE, SimMIM}, discrete tokens\cite{beit, peco}, HOG features\cite{maskfeat}, deep features\cite{pmlr-v162-baevski22a, ibot}, and CLIP\cite{caev2}. A major bottleneck for the industrial applications of MIM is that these models often require large computational resources and substantial pre-training time. As a result, some works have accelerated the process by employing the asymmetric encoder-decoder strategy\cite{MAE, greenmim} or reducing the input patches\cite{chen2022efficient, li2022uniform}. In this work, we leverage scale as a self-supervised signal to learn scale-invariant representation.

\noindent\textbf{Super Resolution} 
Compared to traditional image super resolution methods \cite{gu2012fast, timofte2013anchored, timofte2015a+, michaeli2013nonparametric, he2010single} which are generally model-based, learning based methods have become more popular due to their impressive performance. The convolutional neural network (CNN)-based super-resolution model holds an important position in super-resolution work, with SRCNN\cite{SRCNN} being the first model to utilize CNN in SISR. Since then, many works\cite{EDSR, RCAN, Dai_2019_CVPR, 10.1145/3343031.3351084, Hui_2018_CVPR, LatticeNet} have continuously improved the CNN-based super-resolution model. With the increasing use of Vision Transformer in the field of vision \cite{ViT}, IPT \cite{IPT} has utilized a transformer-based network as a pre-trained model. SwinIR \cite{SwinIR} introduced the Swin Transformer to SISR, while ESRT \cite{ESRT} uses a lightweight CNN backbone to extract deep features and obtain long-range dependency relationships through a lightweight Transformer, combining the CNN and Transformer frameworks. In this study, the HPB module in ESRT \cite{ESRT} was adopted to extract features for resolution recovery since the HPB module aligns well with our prediction head in predicting the original resolution signal from chosen LR patches and remaining HR clues.

\noindent\textbf{VLR/LR recognition} VLR/LR images (or regions of interest) often contain less information content, rendering ineffective feature extraction and classification, thus can intuitively demonstrate the model's learning ability for the features of low-resolution images. In practical applications, we often face low-quality images, which may be low-resolution, noisy, blurry, or have low dynamic ranges.  According to the conclusion in \cite{ImpactOfLow}, even when the degradation in image quality is imperceptible to the naked eye, the performance of deep neural networks will significantly decline. At the same time, the paper also indicates that super-resolution can effectively alleviate the decrease in classification accuracy caused by low resolution. In existing VLR/LR recognition technology, HR information is usually used to improve the classification model, which can be divided into image-level, feature-level, and classifier-level. In terms of image-level technology, Grm et al. \cite{Grm} proposed a cascaded super-resolution network and a set of face recognition models as priors, while Kazemi et al. completed the same task using a multi-scale generative adversarial network \cite{Kazemi}. Researchers have also proposed VLR/LR algorithms that merge HR information at the feature or classifier level \cite{HFR, Aghdam_2019_CVPR_Workshops, Wang_2016_CVPR, studentTeacher}. In the work of \cite{DeriveNet}, effective class boundaries were learned through joint training of two losses, and a new data augmentation based on a multi-resolution pyramid was used for training. In this work, the combination of super-resolution and the MIM model has made our model more robust for low-resolution image tasks. Therefore, we demonstrate the scale invariance ability of our model through the VLR/LR recognition task.

\section{Method}
\subsection{A Revisit of MIM}
The MIM framework involves masking parts of the input image and then predicting their information based on the unmasked observation. It consists of four main components: Masking strategy, encoder architecture, prediction head, and prediction target. The masking strategy determines how to select and mask an area. The encoder architecture extracts a latent feature representation for the masked image, which is then used to predict the original signal within the masked area. The prediction head is applied to the latent feature representation to generate one form of the original signals within the masked area. The prediction target defines the form of the original signals to be predicted.


Both our SRMAE and MAE belong to the MIM framework. Our SRMAE is a modification of MAE's structure, so based on the MIM framework, we will first highlight the similarities and differences between our approach and the well-known MAE. MAE treats the original pixel intensity as the prediction target by reconstructing it, while we use scale as our target by recovering the resolution. The masking strategy in MAE samples a subset of patches and remove the remaining ones, sending the remaining visible patches to the encoder for learning latent representation. Our SRMAE gains low-resolution signals through downsampling and utilizes these removed patches in MAE, which will be explained in more detail later. The encoder used in MAE is a ViT and applied only on sampled patches, which is also the structure we adopted. The prediction head in MAE is a lightweight decoder used for pre-training to perform the image reconstruction task. Our prediction head design differs from that of MAE, and we will elaborate on this in the following section.


\subsection{SRMAE}
Directly applying the original pixel intesity as self-surpervised signal by reconstructing would not be suitable for images with varying scales. We introduce scale as self-supervised signal to utilize the image information at different resolutions. The overall pipeline of SRMAE is shown in Figure \ref{fig:SRMAE_model}.

Applying the original pixel intensity as a self-supervised signal by reconstruction would be unsuitable for images with varying scales. To account for such images, we introduce scale as a self-supervised signal to utilize the information at different resolutions. The overall pipeline of our SRMAE is illustrated in Figure \ref{fig:SRMAE_model}.

\noindent\textbf{Instantiations}
Our work leverages this process to train a novel MIM model, using scale as the self-supervision signal. An example will illustrate the main structure and training process of our model. Given an input image $x$, we first perform the encoder stage. We divide $x$ into non-overlapping image patches, represented by $x^p$. At the same time, we downsample and patchify $x$ to obtain a low-resolution $x_l$. We then perform a masking operation, selecting a random mask $M$ and replacing the corresponding parts of the patches with $[MASK]$, obtaining $x^p_h = x^p\odot  M + [MASK] \odot (1-M)$. $x^p_h$ is a learnable embedding indicating masked patches. We then add positional information to $x^p_h$ and learn the image's latent representation $x_e$ through the Vision Transformer. Then, in the prediction head, we concatenate $x_l$ and $x_e$ based on their corresponding positional information. This replaces the patches marked with $[MASK]$ with the corresponding patches from $x_l$. We preliminarily extract information that favors super-resolution from the concatenated patches sequence using a replaceable super-resolution module. Subsequently, we apply a lightweight Vision Transformer\cite{ViT} that recovers the resolution, resulting in $y^p$. The recovery loss is obtained by calculating the mean squared error (MSE) between $y^p$ and the original image patches $x^p$, $L_{rec} = ||(y^p - x^p) \odot (1-M)||_2^2$. Only the loss generated by the parts replaced by $x_l$ is calculated\cite{MAE}.

During the pre-training stage, we use scale as the self-supervised learning signal, combined with information from input images at different scales to learn the deep cross-resolution representation of images. After pre-training, we abandon the prediction head and only use the encoder for downstream visual tasks under different resolution conditions.

\noindent\textbf{Downsampling}
To utilize scale as a self-supervised signal for training, one must obtain images with resolutions that differ from the original image. Additionally, in order to maintain the positional information of images, it is required to stitch together images with different resolutions to their initial positions once resolution changes have been made. As such, the downsampling process involves two steps: image downsampling and dimensions adaptation.

As show in bottom part of \ref{fig:SRMAE_model}. The first step is to perform a downsampling process. In order to fully utilize the latest advanced super-resolution structure subsequently when restoring the image's resolution, we referred to some papers on super-resolution\cite{SRCNN, 9416983,ESRT, IPT} where the frequently used scaling factors are 2, 3, and 4. Therefore, in this study, we applied these ubiquitous scaling factors to take full advantage of this advanced technology, which is widely recognized in the literature for its effectiveness. For example, when using a scaling factor of 4, we downscaled the original $224\times224$ image to 1/4 of its resolution, resulting in a $56\times56$ image. In the second step, we concatenated the high-resolution patches with the low-resolution patches. Since the dimensions of HR and LR images do not match, we used the nearest neighbor interpolation method to resize the LR image to $224\times224$ dimensions. After that, we divided this interpolated image into non-overlapping patches of size $16\times16$.

\begin{figure}[t]
\begin{center}
   \includegraphics[width=1\linewidth]{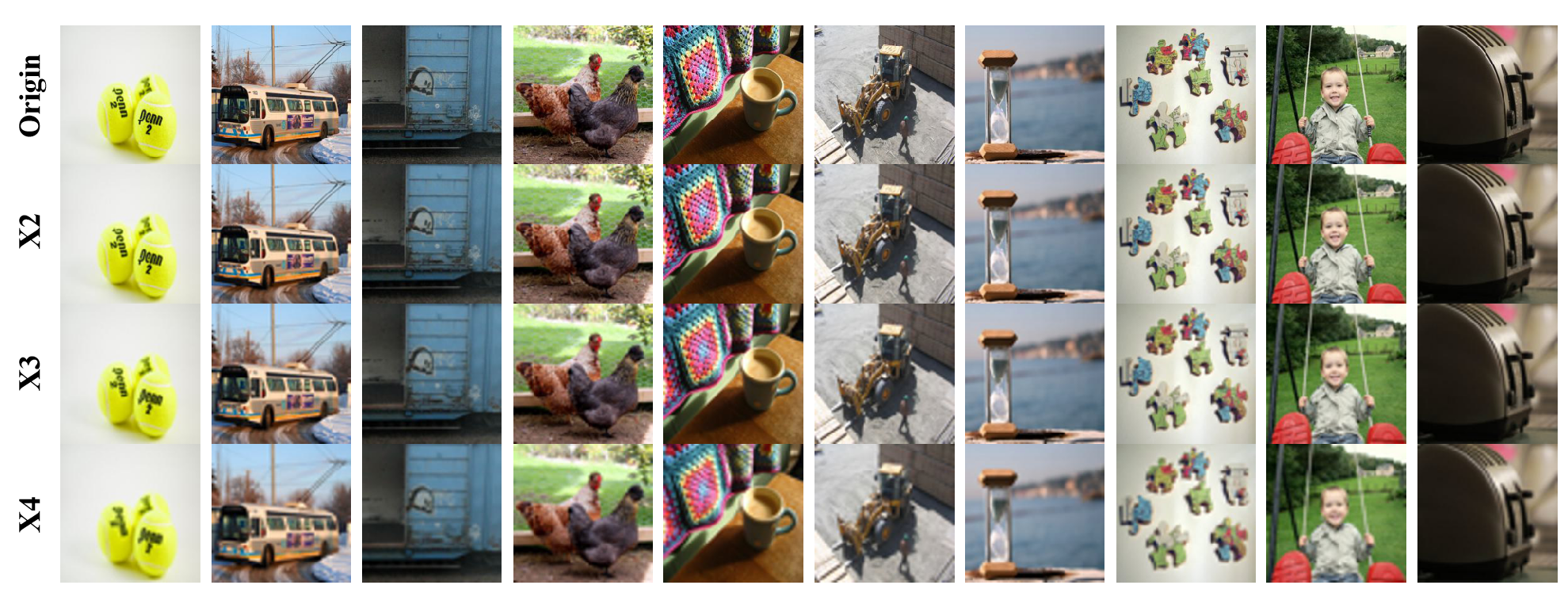}
\end{center}
   \caption{Downsampling images produces visualizations with scale ratios of 1/2, 1/3, and 1/4, respectively. The $\times$2, $\times$3, and $\times$4 notations indicate that the resolution must be expanded by two, three, and four times, respectively, to restore it to its original resolution. Consequently, the resolution is reduced to half, two-thirds, or three-quarters during the downsampling process.}
\label{fig:long}
\label{fig:onecol}
\end{figure}

\noindent\textbf{Encoder}
Only visible patches  $x_p = \{x_p^i|M^i = 1\}$ are fed to the encoder following\cite{MAE} and mapped to the high resolution patches $x_e$ across a stack of transformer blocks. The operation of encoder is based on self-attention. In this paper, we utilize ViT-Base to form the encoder.

\noindent\textbf{Prediction head}
Our study aimed to create a prediction head that could predict resolution recovery in conjunction with super-resolution technology. In contrast to the prediction head design of MAE, our SRMAE model's input for the prediction head consists of a patch sequence composed of high-resolution patches $x_e = \{x_e^i|M^i = 1\}$ output by the encoder phase and the low-resolution patches $x_l = \{x_l^i|M^i = 0\}$. We altered the input of the prediction head to enable resolution restoration instead of image reconstruction by using low-resolution image patches in place of [MASK] training patches utilized in MAE.

The prediction head, illustrated in \ref{fig:SRMAE_model}, consists of a High Preserving Block (HPB) module and a lightweight Vision Transformer (ViT). To enhance the resolution recovery ability of the prediction head for super-resolution tasks, we extracted applicable modules from advanced super-resolution works. The HPB module proposed in ESRT captures image texture details in its own super-resolution model structure, making it suitable for feature extraction from both high-resolution and low-resolution patches. As such, we used it as the first processing module in our prediction head. The implementation of a transformer in IPT for super-resolution tasks demonstrates that the lightweight ViT used in our prediction head also possesses super-resolution ability. As a result, by combining effective modules from super-resolution works with the ViT in the prediction head, we achieved resolution recovery of low-resolution patches.

\section{Experiments}
The SRMAE framework is designed for learning scale-invariant representations. In this section, we conducted extensive experiments to evaluate the effectiveness of the SRMAE framework in learning scale-invariant representations. We evaluated the ability of SRMAE to maintain scale invariance for high-resolution images by conducting experiments on the ImageNet-1K dataset \cite{imagenet1k}. We fine-tuned the entire set of parameters of the classification task using a fine-tuning protocol and reported the results of transfer learning to assess the quality of scale invariance for high-resolution images. In addition, we conducted numeric classification tasks on the SVHN dataset\cite{SVHN} at very low resolutions and low-resolution object recognition tasks on the CK+\cite{CK+}, RAF-DB\cite{RAFDB}, and ExpW\cite{expw} datasets to assess the ability of SRMAE in preserving scale invariance for low-resolution images.
\begin{table}[t!]
\small
\setlength{\tabcolsep}{3pt}
\begin{center}
\begin{tabular}{lcccccc}
\toprule
Methods & Pretraining Data & Epochs & Finetuning Resolution & Pretraining Time (h)& Top-1 Acc.\\
\midrule
scratch\cite{deit} & -  & -  & 224$\times$224 & - & 81.8  \\
$AT^{*}$ \cite{AT} & IN1K & - & - & - & 65.4 \\
$CCKD^{*}$ \cite{CCKD} & IN1K & - & - & - & 67.7 \\
DINO \cite{dino} & IN1K  & 300  & 224$\times$224 &- & 82.8 \\
MoCov3 \cite{mocov3}& IN1K  & 600 & 224$\times$224 &-&   83.2 \\
BEiT \cite{beit} & IN1K+DALLE \cite{ramesh2021zero}  & 800 & 224$\times$224 & $\sim$285 &83.2   \\
MAE \cite{MAE} & IN1K  &400 & 224$\times$224 &$\sim$440&83.1\\
MAE\cite{MAE} & IN1K   &800 & 224$\times$224 &$\sim$880&83.3 \\
MAE\cite{MAE}& IN1K   &1600 & 224$\times$224 &$\sim$1760&83.6 \\
\rowcolor{Gray2}
SRMAE (Ours) & IN1K  &400 & 224$\times$224 &$\sim$896& 82.1\\
\bottomrule
\end{tabular}
\end{center}
\caption{\textbf{Comparison with previous work on IN-1K.} Except for the two methods with the $*$ symbol, all other entries are pre-trained on IN-1K train split and all other models are pre-trained and fine-tuned under 224 x 224 resolution. The pretraining time is estimated on a machine with one Tesla V100-32G GPU, ‘GPU Hours’ is the running time on single GPU. The methods with the $*$ symbol represent the ability to handle low-resolution tasks, and have good performance in low-resolution tasks, but they are not self-supervised models.}
\label{tab:IN-1K}
\end{table}

\subsection{High-quality Images Tasks}
\noindent\textbf{Settings}
The imageNet-1K(IN-1K)\cite{imagenet1k} dataset contains 1.3 million images categorized into 1000 classes for image classification and is divided into training and validation sets. The evaluation protocol is pre-training followed by end-to-end fine-tuning. We utilized vanilla ViT\cite{ViT} base models without modification and pre-trained our models without labels on the IN-1K training set at a resolution of $224^2$. To minimize data augmentation, we used random resized cropping and horizontal flipping. We randomly mask out 75\% of total image patches following MAE\cite{MAE}. For downsampling the images during pre-training, we employed the commonly-used four-fold downsampling strategy used in super-resolution experimentation\cite{SRCNN, ESRT}. We follow the default finetuneing parameters of the MAE\cite{MAE}. For finetuning, we report the classification accuracy on the IN-1K validation set of the finetuned SRMAE encoders.


\noindent\textbf{Classification on ImageNet-1K} 
We 	present the accuracy of SRMAE on Table \ref{tab:IN-1K} and conduct comparisions with previous mask autoencoding methods. For fair comparison, we estimate the pre-training duration of each model on the same machine with one Tesla V100-32G GPU. We report the running time on single GPU, denoted as 'Pretraining Time'. We pre-trained SRMAE for 400 epochs, obtaining a top-1 accuracy of 82.1\%, which is 0.3\% higher than from-scratch. Compared with the MAE pretrained for 400 epochs, our finetuning accuracy is 1\% lower with the same number of pretraining epochs. Our finetuning accuracy is 1.2\% lower than that of the MAE pretrained for 800 epochs, despite the close pretraining time. CCKD\cite{CCKD} is a comparative study in subsequent LR recognition experiments, and it also reports the top-1 accuracy on ImageNet-1K. Compared with CCKD, our top-1 accuracy improved by 14.4\%, far exceeding their results. We believe that the use of scale signal leads to s subpar from-scratch and fine-tune performance, and leave further investigations of this phenomenon to future work. We note, however, that our method still improves over these handling low-resolution models by as large a margin as other methods. Our method still achieves significant improvement compared to previously used models (CCKD\cite{CCKD}, AT\cite{AT}) for low-resolution tasks.


\begin{figure}[t]
\begin{center}
   \includegraphics[width=1\linewidth]{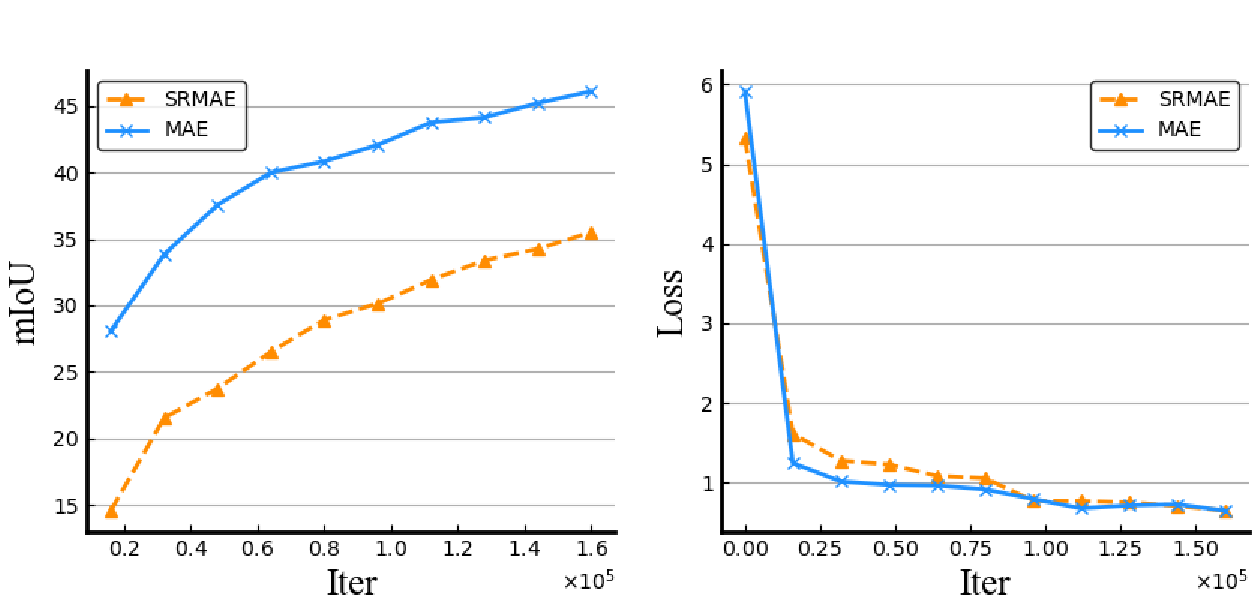}
\end{center}
   \caption{\textbf{Semantic segmentation on ADE20K} The left chart shows mIoU scores at different training iterations. The right chart presents loss values collected at various training iterations. We compared the transfer learning performance of MAE and SRMAE in a semantic segmentation task.}
\label{fig:linegraph}
\end{figure}

\noindent\textbf{Semantic segmentation on ADE20K} 
We utilized UperNet \cite{UPerNet} and followed the semantic segmentation code of \cite{beit, MAE}. Our SRMAE's mIoU, as presented in Figure \ref{fig:linegraph}, was 14.55 at the beginning of training and improved to 35.5 at the end, a rise of 20.95 points. To compare our results, we replicated MAE's semantic segmentation experiment, which yielded an initial mIoU score of 28.05, which improved to 46.1 at the end, a rise of 18.05 points. Although our transfer learning ability was inferior to MAE, our training improvement was 2.9 points higher. We evaluated the loss results and observed that the loss values were consistently within similar levels, as shown on the right side of Figure \ref{fig:linegraph}. However, on the left-hand side, a noticeable gap in mIoU persisted. Because we based our code on MAE's implementation, we speculate that our model modifications made the initialization of other segments irregular, leading to unsatisfactory mIoU scores.

\subsection{Low-quality Images Tasks}
\noindent\textbf{Datasets} The SRMAE has been evaluated on two different tasks: VLR digit classification and LR facial expression classification. The first task utilized one dataset, whereas the second task used three datasets to evaluate the scale invariance ability of SRMAE. Details for each benchmark datasets ars as below:


(i) SVHN\cite{SVHN}: We employed the Street View House Numbers (SVHN) dataset to assess the efficacy of SRMAE in classifying very low-resolution digits. This dataset comprises 0-9 digit images that were captured from natural scenes in the real world, at a 32$\times$32 resolution. It consists of 73,257 images for training and 26,031 images for testing. Consistent with established protocols set forth by Wang, et al. \cite{Wang_2016_CVPR}, the test dataset was created via subsampling the images by a factor of 2, resulting in an 8$\times$8 resolution that represents the lowest resolution among the four datasets.

(ii) CK+\cite{CK+}: The CK+ dataset contains 593 image sequences, each describing the transition process from a neutral expression to an emotional peak. We followed the protocol of \cite{CKset} and used six prototype expressions (anger, disgust, fear, happiness, sadness, and surprise) in the training and testing process. In 327 image sequences, we selected the first frame as the neutral expression. Therefore, we obtained a total of seven expressions, with 1254 facial expression images, including the background, at a resolution of 640 $\times$ 490 pixels. In the experiment, we converted the images to 100 $\times$ 100 pixels for low-resolution testing.

(iii) RAF-DB\cite{RAFDB}: The RAF-DB database contains 12,271 training images and 3,068 testing images, with each picture being 100 $\times$ 100 pixels and using seven emotion labels including anger, disgust, fear, happiness, sadness, surprise, and neutral.

(iv) ExpW\cite{expw}: The ExpW database is one of the largest facial expression databases, containing 91,793 network images. The facial area of these images ranges from 23$\times$23 pixels to 180$\times$180 pixels. We use the dlib library to extract the facial area from these images. To meet the experimental requirements, images with a facial area smaller than 80$\times$80 pixels were filtered out, while those equal to or greater than 80$\times$80 pixels were resized to 100$\times$100 pixels. In the end, a total of 31,127 processed images were obtained.

\begin{table}[t!]
\small
\setlength{\tabcolsep}{3pt}
\begin{center}
\begin{tabular}{lcccccc}
\toprule
Algorithm & Top-1 Acc. & Top-5 Acc.\\
\midrule
CNN(VLR)(2016)~\cite{cnnlr}              & 45.29 & 66.78\\
RPC Nets(2016)~\cite{cnnlr}                      & 56.98 & 70.82   \\
BSICNN(2018)~\cite{SICNN}                       & 81.51  & 96.77 \\
CapsNet(VLR)(2019)~\cite{CapsNet}                         & 79.19 & 88.89     \\
DirectCapsNet(2019)~\cite{CapsNet}                     & 84.51 &  91.20 \\
CSRIP(2020)~\cite{Grm}                     & 84.62 & 97.32   \\
DeriveNet(2022)~\cite{DeriveNet}           & 87.86 & 97.18 \\
\rowcolor{Gray2}
MAE\cite{MAE} & \textbf{89.10} &  \textbf{98.38}\\ 
\rowcolor{Gray2}
SRMAE (Ours) & \textbf{89.14} &  \textbf{98.40}\\ 
\bottomrule
\end{tabular}
\end{center}
\caption{\textbf{Top-1 and Top-5 Accuracy(\%) on SVHN Dataset\cite{SVHN} for VLR Digit Recognition(8 $\times$ 8)}}
\label{tab:SVHN}
\end{table}

\textbf{VLR digit classification on SVHN dataset} As the SVHN dataset contains sufficient data, we pretrain our SRMAE on SVHN dataset without label followed by end-to-end fine-tuning on SVHN dataset. Table \ref{tab:SVHN} presents the top-1 and top-5 classification accuracies on the SVHN dataset. The SRMAE model achieved a top-1 classification accuracy of 89.14\% and a top-5 classification accuracy of 98.40\%, demonstrating an improvement of over 1\% from the state-of-the-art result.We used MAE to conduct the same experiment, and the final results showed that the top-1 classification accuracy was 89.10\%, while the top-5 classification accuracy was 98.38\%. Our results show an improvement compared with those reported.

The improved performance demonstrates that using scale as a self-supervised signal can enhance the ability of the model to recognize extremely low-resolution images(e.g., digits) and achieve advanced levels.

\begin{table}[t!]
\small
\setlength{\tabcolsep}{3pt}
\begin{center}
\begin{tabular}{lcccccc}
\toprule
\multirow{2}{*}{Algorithm} & \multicolumn{2}{c}{CK+\cite{CK+}} & \multicolumn{2}{c}{RAF-DB \cite{RAFDB}} & \multicolumn{2}{c}{ExpW\cite{expw}} \\
 & Extra Data & Top-1 Acc. & Extra Data & Top-1 Acc. & Extra Data & Top-1 Acc. \\
\midrule
Ge et al.(2020)~\cite{studentTeacher}          &  -  & 89.92 &  -   & 85.98 &  -  & 63.47\\
Massoil et al(2020).~\cite{massoli2020cross}   &  -  & 91.13 &  -  & 85.07&  -  & 62.96 \\
DGKD(2021)~\cite{DGKD}                         &  -  & 91.94 &  -  & 85.98 &  - & 64.54\\
DirectCapsNet(2019)~\cite{DirectCapsNet}       &  -    & 91.53  &  - & 86.05&  -  & 63.97\\
MSAD(2021)~\cite{MSAD}                         &  -    & 89.11 &  -  & 84.39&  -  & 63.36   \\
Fitnet(2014)~\cite{fitnets}                    &  -    & 92.74 &  -  & 85.85&  -  & 64.97 \\
AT(2017)~\cite{AT}                             &  -    & 91.94&  -   & 86.08&  -  & 65.09   \\
PKT(2018)~\cite{PKT}                           &  -    & 92.74 &  -  & 86.44 &  - & 64.27\\
AFDS(2020)~\cite{AFDS}                         &  -    & 91.13&  -   & 86.02 &  - & 64.65 \\
AB(2022)~\cite{AB}                             &  -    & 94.36 &  -  & 86.67&  -  & 64.37 \\
VID(2022)~\cite{VID}                           &  -    & 92.34 &  -  & 86.08&  -  & 64.40 \\
FAKD~\cite{FAKD}                               &  -    & 77.82 &  -  & 86.51 &  - & 64.08\\
OFD\cite{OFD}                                  &  -    & 93.15&  -   & 85.82 &  - & 63.55\\
FT\cite{FT}                                    &  -    & 93.95 &  -  & 86.47 &  - & 64.29 \\
VKD\cite{VKD}                                  &  -    & 93.15 &  -  & 85.20 &  - & 63.42 \\
EKD\cite{EKD}                                  &  -    & 90.32&  -   & 84.78 &  - & 63.19 \\
CCKD\cite{CCKD}                                &  -    & 91.94 &  -  & 86.12 &  - & 64.90\\
RKD\cite{RKD}                                  &  -    & 20.16  &  - & 11.41 &  - & 11.20\\
RKD*\cite{RKD}                                 &  -    & 92.34  &  - & 85.76 &  - & 63.28\\
SP\cite{SP}                                    &  -    & 94.76  &  - & 86.96 &  - & 64.12 \\
FMD\cite{FMD}                                  &  -    & \textbf{95.57} &  - &\textbf{87.13}& - & 65.37 \\
\rowcolor{Gray2}
MAE\cite{MAE} & IN1K & $\text{95.04}^*$ & IN1K  &  $\text{85.72}^*$ & - & 74.53\\ 
\rowcolor{Gray2}
SRMAE (Ours) & IN1K & $\text{94.21}^*$ & IN1K & $\text{82.70}^*$ & - & \textbf{74.84}\\ 
\bottomrule
\end{tabular}
\end{center}
\caption{\textbf{Compared to previous work in low-resolution object recognition tasks} Top-1 accuracy results for low-resolution object recognition obtained on CK+, RAF-DB, and ExpW datasets. Best results are bold-faced.* represents that the pre-training of the MAE model was conducted using the IN-1K dataset for 1600 epochs, while the pre-training of the SRMAE model was carried out using the IN-1K dataset for 400 epochs.}
\label{tab:lowobject}
\end{table}

\textbf{LR facial expression classification on CK+ dataset}: In the experimental setup, we discovered that the CK+ dataset had limited data. Thus, training the model using CK+ dataset alone may not produce sophisticated results. We employed the model that was pre-trained for 400 epochs on the ImageNet-1K\cite{imagenet1k} dataset and fine-tuned it on the CK+ dataset. According to the results presented in \ref{tab:lowobject}, our model attained a top-1 accuracy of 94.21\% on this dataset, which is within 1.4\% of the state-of-the-art performance. We conducted an unfair comparison with MAE, due to the difference in the pre-training process. MAE employed a ViT-B model that was pre-trained for 1600 epochs and fine-tuned it for the downstream task, resulting in a top-1 accuracy of 95.04\%.

\textbf{LR Facial Expression Classification on RAF-DB dataset}: The size of the RAF-DB dataset is still relatively small. We also use the model that was pre-trained for 400 epochs on the ImageNet-1K dataset. We fine-tuned it on the RAF-DB dataset. According to the results presented in \ref{tab:lowobject}, the model's performance on the RAF-DB dataset was found to be 5\% lower than that of the state-of-the-art models. We also conducted an unfair comparison with MAE dut to MAE employed a model that was pre-trained for 1600 epochs. The results in \ref{tab:lowobject} demonstrated that MAE yielded good results on the RAF-DB dataset. Therefore, we attribute our relatively lower performance to the inadequate pre-training of our model. If more comprehensive pre-training is undertaken, we anticipate that our results may significantly improve.

\textbf{LR Facial Expression Classification on ExpW dataset}: The ExpW dataset is one of the largest databases consisting of facial expressions. Hence, we forewent the use of supplementary datasets \cite{imagenet1k} and exclusively employed the ExpW dataset for pre-training, followed by fine-tuning during downstream tasks. Our model yielded a top-1 classification accuracy of 74.84\%, as per the results presented in \ref{tab:lowobject}. This accuracy is 9.48\% higher than that of the previous state-of-the-art model, indicative of a significant improvement in our methodology. Additionally, we conducted a fair comparison with MAE on this dataset, without using any extra pre-training datasets. Our experimental results revealed that our model achieved a top-1 accuracy of 74.53\% on this dataset, which is marginally below that of SRMAE by 0.3\%.

SRMAE exhibits strong robustness in low-resolution tasks, particularly in larger datasets where the advantages of self-supervised learning can be fully utilized, leading to excellent performance. Moreover, SRMAE improves the MIM model's learning of low-resolution image features under similar experimental conditions by using scale as the self-supervisory signal, which renders it more suitable for low-resolution tasks compared to MAE.

\section{Conclusion and Discussion}
We present a robust and effective MIM framework, named SRMAE, that is capable of learning  scale-invariant deep representation and apply it to visual tasks of different scales. TOur method leverages scale as a self-supervised signal to facilitate resolution restoration, which replaces image reconstruction and adapts easily to scale-varied tasks. We have achieved extensive results in tasks such as high-resolution image classification and low-resolution object recognition. To the best of our knowledge, this is the first model that achieves close to SOTA results for high-resolution image classification and low-resolution object recognition.


A natural extension in the future would involve using more modules that can enhance the super-resolution ability to further improve performance.

{\small
\printbibliography
}

\end{document}